\documentclass[letterpaper, 10 pt, conference]{ieeeconf}  
\IEEEoverridecommandlockouts                              
\usepackage{graphics} 
\usepackage{epsfig} 
\usepackage{mathptmx} 
\usepackage{times} 
\usepackage{amsmath} 
\usepackage{amssymb}  
\usepackage{siunitx}

\title{\LARGE \bf
Meta-Ori: monolithic meta-origami for nonlinear inflatable soft actuators}

\author{Hugo de Souza Oliveira, Xin Li, Johannes Frey, Edoardo Milana*
\thanks{All authors are with livMatS@FIT, University of Freiburg, Freiburg, Germany and the Department of Microsystems Engineering - IMTEK, University of Freiburg, Freiburg, Germany}
\thanks{*Corresponding author:
       {\tt\small milana@imtek.de}}%
}

\begin{document}

\maketitle
\thispagestyle{empty}
\pagestyle{empty}

\begin{abstract}

The nonlinear mechanical response of soft materials and slender structures is purposefully harnessed to program functions by design in soft robotic actuators, such as sequencing, amplified response, fast energy release, etc. However, typical designs of nonlinear actuators - e.g. balloons, inverted membranes, springs - have limited design parameters space and complex fabrication processes, hindering the achievement of more elaborated functions. Mechanical metamaterials, on the other hand, have very large design parameter spaces, which allow fine-tuning of nonlinear behaviours. In this work, we present a novel approach to fabricate nonlinear inflatables based on metamaterials and origami (Meta-Ori) as monolithic parts that can be fully 3D printed via Fused Deposition Modeling (FDM) using thermoplastic polyurethane (TPU) commercial filaments. Our design consists of a metamaterial shell with cylindrical topology and nonlinear mechanical response combined with a Kresling origami inflatable acting as a pneumatic transmitter. We develop and release a design tool in the visual programming language Grasshopper to interactively design our Meta-Ori. We characterize the mechanical response of the metashell and the origami, and the nonlinear pressure-volume curve of the Meta-Ori inflatable and, lastly, we demonstrate the actuation sequencing of a bi-segment monolithic Meta-Ori soft actuator. 

\end{abstract}


\section{Introduction}

During the last decade, soft robotics has attracted significant attention in various sectors, including medical \cite{cianchetti2018biomedical}, agricultural \cite{armanini2024soft}, and maritime \cite{aracri2021soft}. This growing interest derives from the inherent advantages of soft robots, caused by their compliant structures and materials, which make them uniquely suited for applications where attributes such as safety and adaptability are critical \cite{DellaSantina2020}.

As conventional electronics and control algorithms are challenging to implement in soft robots, due to materials (rigid PCBs) and modeling (finite number of degrees of freedom) incompatibilities, new approaches are explored to craft autonomous behaviors in soft robotics. Following bioinspired paradigms, a notable approach is \textit{physical intelligence} \cite{sitti2021physical}, where simple control rules and feedback loops are implemented in the physical characteristics of the materials and structures that constitute the soft robots.

\begin{figure}[htbp]
    \centering
    \includegraphics[width=1\linewidth]{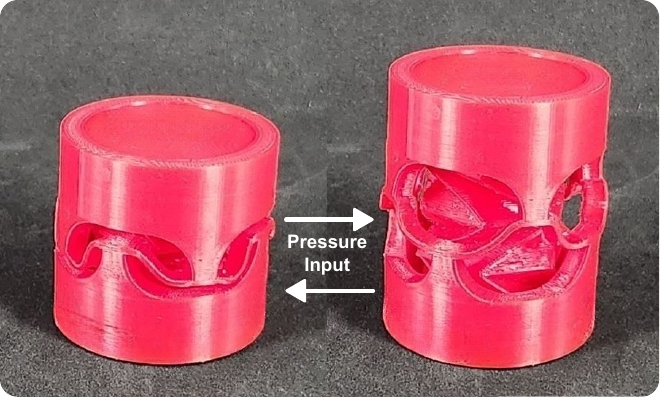}
    \caption{Meta-Ori bistable inflatable actuator in fully open and closed configurations.}
    \label{FIG-1}
\end{figure}

Different design strategies have been explored to enable such capabilities in soft robots, ranging from the responsiveness of smart materials \cite{qian2019artificial} to the nonlinear mechanics of slender soft material structures \cite{pal2021exploiting}. The latter is particularly interesting because it depends mostly on the geometry of the structure rather than the chemical constituents of the material. Nonlinear mechanical effects, such as buckling and snap-through instabilities, have been used to create fluidic logic \cite{rothemund2018soft}, actuation sequencing \cite{milana2018design,van2023nonlinear} and fast motions \cite{tang2020leveraging}, harnessing structures like balloons \cite{ben2020single,milana2022morphological}, buckled membranes \cite{gorissen2020inflatable} and multi-stable origami \cite{melancon2022inflatable}. However, these structures have limited design parameters space, which in turn hinders the tunability of their mechanical response. Moreover, they undergo fundamental mechanical limitations, e.g. balloons tend to burst after a limited amount of snap-through events due to extreme stretches, and membranes and origami have limited snap-through displacements.

\begin{figure*}
    \centering
    \includegraphics[width=1.0\linewidth]{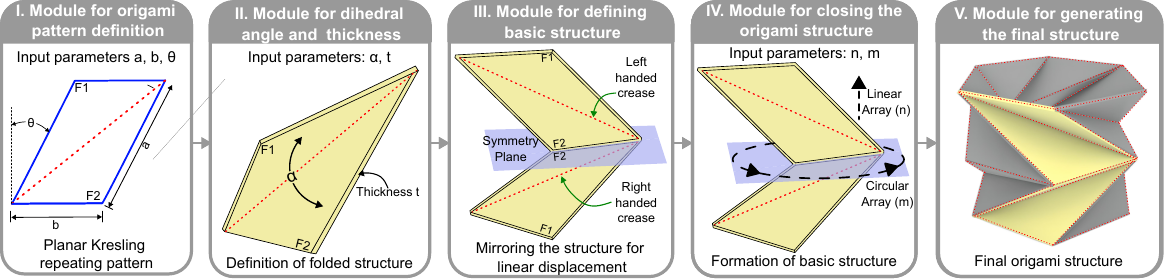}
    \caption{Sequential steps to generate the Kresling origami structure implemented in the Grasshopper visual code.}
    \label{FIG-2}
\end{figure*}

A design approach that dramatically increases the design space lies in the use of mechanical metamaterials, that is, engineered structures that exhibit behaviors that are not typically found in natural materials due to their internal artificial architecture \cite{bertoldi2017flexible}. Snapping mechanical metamaterials can have cylindrical topologies (metashells) \cite{yang_1d_2020}, which make them ideal for fluidic actuation. Moreover, the overall structure can undergo substantial elongation (up to 100\%) according to the unit-cell geometry. However, creating snapping inflatable actuators using metashells is challenging because an enclosed cavity must be added to transmit the fluid pressure to the structure. A simple enclosed cavity like a rubber balloon would add axial stiffness to the structure if the material is as soft as the shell, linearising the response and canceling out the snapping effect. If the material is too soft and loop strains are not constrained, the balloon will bulge out of the shell.  

In previous work, we used a conventional rigid single-acting piston (without a spring) to actuate a metashell \cite{navale2024bistable}, losing however the softness of the actuator. A promising solution for entirely soft inflatables was presented by Rahman et al. \cite{rahman2023zero} that made a multistable inflatable actuating a metashell with a bellow pressure transmitter made of inextensible materials, which has essentially a near-zero axial stiffness and high circumferential stiffness due to the inextensibility of the material. However, the fabrication method requires many manual gluing steps and the final manual assembling of the transmitter to the metashell.

In this paper, we present a novel nonlinear inflatable actuator, named Meta-Ori, which is composed of a snapping metashell that is actuated by the geometric transformation of a hexagon Kresling-based inflatable origami. The origami's elongation, induced by fluidic pressure, triggers the metashell to snap. The foldable mechanism of the Kresling origami introduces low axial stiffness, behaving more like a bellow than a balloon, and allows us to use it as a pressure transmitter. Most importantly, this integration allows for the fabrication of the combined system in a single process using 3D printing, eliminating many intermediate steps such as molding, gluing, and manual assembly. Our design achieves larger strokes compared to balloons and other snapping origami. Fig. \ref{FIG-1} shows the Meta-Ori inflatable in its open and closed configuration with the origami-based actuator inside. 


Furthermore, designing the origami transmitters presents several challenges, as many geometric parameters must be considered, requiring optimizations or iterative attempts to find the most adequate structure. Unfortunately, most of the software dedicated to designing origami is built considering that the origami will be constructed using sheet-like materials such as paper, which means they are not very suitable to be used as inflatables. Therefore, we also developed a Grasshopper visual code that generates the origami structure based on the input geometric parameters. This tool facilitates the seamless integration of the origami structure with the metashell.

\section{Design \& Fabrication}

\subsection{Grasshopper visual code for origami inflatable design}

The visual code is developed using the Grasshopper plugin from Rhinoceros 3D version 8, as it offers a parametric design flexibility not commonly found in traditional 3D CAD software. Grasshopper’s visual programming interface allows for real-time adjustments to geometric parameters, enabling efficient exploration of the design space. This design flexibility is crucial when working with complex geometries such as origami structures, in which small changes in the geometric parameters can significantly affect the actuator’s kinematics, the mechanical behavior under operation, and its integration with other structures.

Unlike traditional origami design and fabrication, in which creases are defined on sheet-like materials and folded into a final shape, 3D printing offers advantages, such as no additional assembly and gluing of edges, making the fabrication process more straightforward and simple.

Fig. \ref{FIG-2} illustrates the sequential steps followed by the visual code to generate the Kresling origami structure. Each box represents a module that processes input parameters to construct each stage of the origami's definition. In Fig. \ref{FIG-2}.I, the module defines the planar origami pattern, where blue and red dotted lines represent mountain and valley folds, respectively. Key parameters such as the edge lengths $a$ and $b$ as well as the inclination angle $\theta$ are input to generate spatial points that meet geometric constraints, forming faces F1 and F2 of the Kresling pattern. 

Fig. \ref{FIG-2}.II shows the module for defining the dihedral angle $\alpha$ (the initial angle of the origami) and the thickness $t$ of the faces F1 and F2. The intersecting edges represent right-handed creases. Once the folded structure is formed, the basic structure is completed (Fig. \ref{FIG-2}.III). A symmetry plane is used to mirror the folded structure, creating an inverted crease direction (left-handed crease). This configuration of opposing crease directions ensures that the top and bottom parts of the basic structure remain aligned during vertical folding and unfolding, as the rotation is concentrated along the intersecting edges in the mirror plane.

In Fig. \ref{FIG-2}.IV, the module for closing and finishing the structure generates the basic unit in both a circular and vertical array. The circular array is centered on a hexagon with side length $b$, as defined in Fig. \ref{FIG-2}.I, while the vertical array is arranged according to the specific structure where the origami will be integrated. The minimal configuration required to ensure alignment of the top and bottom parts is shown in Fig. \ref{FIG-2}.V.

\subsection{Metashell design}

The snapping metashell is a metamaterial with cylindrical topology designed to exhibit a non-monotonic force-displacement relationship under tensile loading, which leads to a snap-through instability at a specific force threshold, allowing it to transition between two different configurations.

The metashell design is based on a unit cell with double-clamped curved beams, defined by the relation $y(x) = \frac{h}{2}[1 - \cos(2\pi x / l)]$ within the range $[-l/2, 0]$, following the approach described in \cite{yang_1d_2020}. We modified the unit cell design by adding a gradual increase of the support beam width $c$ to diminish out-of-plane deformations of the support beams and increase the overall stiffness of the metashell. To achieve symmetry, a 1x4 array of unit cells was selected, with the origami actuator centrally positioned to align along the metashell’s deformation axis.

\begin{figure}
    \centering
    \includegraphics[width=1\linewidth]{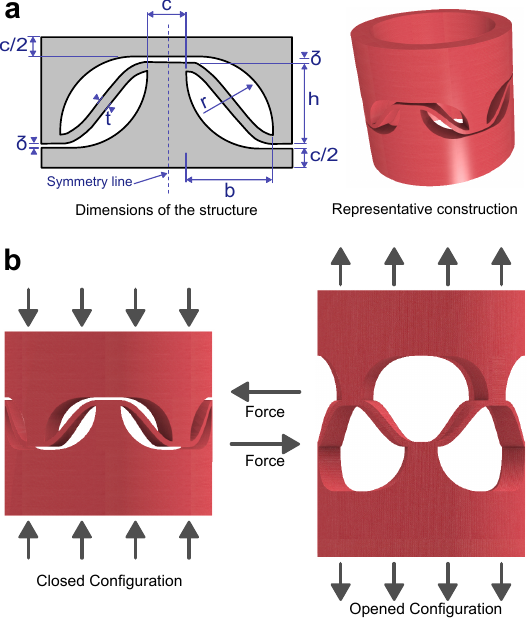}
    \caption{Geometric definition of the metamaterial: a) Sketch and dimensions of the bistable structure with representative construction; b) Metashell in open and closed configurations.}
    \label{FIG-4}
\end{figure}

Fig. \ref{FIG-4}a provides an unrolled view of the unit cell to illustrate its dimensions and details in a planar format, where $c=\SI{12.50}{\milli \meter}$, $l=\SI{22.50}{\milli \meter}$, $t=\SI{1.25}{\milli \meter}$, $h=\SI{9.40}{\milli \meter}$, $r=\SI{7.60}{\milli \meter}$, and $\delta=\SI{0.63}{\milli \meter}$. An additional wall height of \SI{12.5}{\milli \meter} above and below the structure improves mechanical stability, as demonstrated in Fig. \ref{FIG-4}b, which shows the metashell in both open and closed configurations. The overall dimensions of the metashell are \SI{42.50}{\milli \meter} in height, \SI{46.25}{\milli \meter} in diameter, and \SI{36.25}{\milli \meter}, as inner diameter.

\subsection{Integration of the metashell with the origami inflatable}

\begin{figure}[b]
    \centering
    \includegraphics[width=1\linewidth]{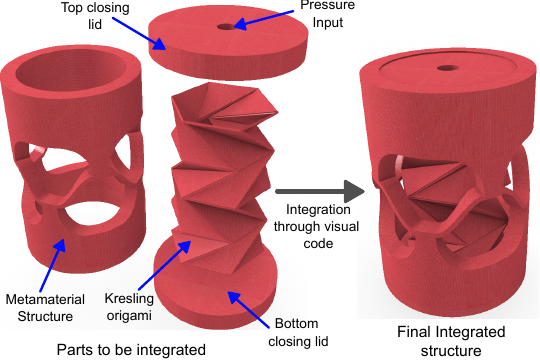}
    \caption{Description of the integration process to generate the Meta-Ori monolithic structure. The Kresling origami is placed in the middle of the metashell and sealed between two lids.}
    \label{FIG-5}
\end{figure}

The 3D solid file of the metashell (designed using SolidWorks) is imported into Rhinoceros 3D and merged with the origami generated by the Grasshopper code. The two structures are aligned to ensure that the actuation movement of the origami is parallel to the mechanical state transition of the metamaterial, as shown in Fig. \ref{FIG-5}. To fix the origami within the metamaterial, additional supports are incorporated, enclosing the origami structure to form the inflatable cavity. The bottom support includes a central hole to allow for pressure input. In this design, the metashell and the origami are similar to two springs in parallel, which have the same displacement but different contributions to the elastic response, $F_{el} = F_{meta} + F_{ori}$. The rationale for the design is that the mechanics of the metashell dictate the overall nonlinear response of the Meta-Ori, where the origami acts mostly as a pneumatic transmitter.

\subsection{Fabrication process} 

All the structures fabricated in this work - metashell, origami, and the combined Meta-Ori - are printed in thermoplastic polyurethane (TPU) with a Shore hardness of 85A (NinjaFlex, NinjaTek Inc.) using a multi-material fused deposition modeling (FDM) 3D printer (Prusa XL). Butenediol vinyl alcohol co-polymer (BVOH from Verbatim Inc.) is used as the supporting material, which is later removed by sonicating the printed structures in water for \SI{4}{\hour}. The final Meta-Ori print is shown in Fig. \ref{FIG-1}. 

To ensure high print quality and well-defined faces and edges of the origami, a low nozzle speed is used, specifically \SI{10}{\milli \meter \per \second} with an acceleration of \SI{10}{\milli \meter \per \second \squared} for the external perimeters. This helps to maintain print integrity and minimizes defects in the origami structure. For the remaining sections of the Meta-Ori structure, standard TPU printing speeds are applied. We vary the percentage of grid infill between 60\% and 99\%, where the infill density only has an impact on the stiffness of the support beams and walls.

To seal the origami cavity and minimize fluid leakage, Ecoflex 00-30 is poured into the Meta-Ori through the pressure input hole. The Meta-Ori is then rotated for \SI{20}{\second} using a Kurabo mixer machine at \SI{10}{rps}, ensuring an even distribution of an Ecoflex layer on the inner surfaces of the origami to seal any potential gaps and prevent leaks.

\subsection{Experimental setup}
\subsubsection{Mechanical test} The mechanical response of the metashell, the origami, and the integrated Meta-Ori is tested using a uniaxial testing machine (zwickiLine from ZwickRoell, 50 N load cell). The test consists of 5 cycles of compression and tension with a controlled displacement rate of 60 mm/min.
\subsubsection{Pressure-volume test} The quasi-static pressure-volume curves of inflatable Meta-Oris are characterized using a motorized syringe pump (Harvard Apparatus PHD ULTRA) to control volume flow and a pressure transducer (KELLER PAA-21Y) to measure the cavity pressure and a video camera to capture actuator deformation. The data is acquired through an NI DAQ USB-6212 and analyzed in a custom Python environment. Each test consists of 10 cycles of inflation-deflation at a fixed volume flow of 60 mL/min.

\section{Results \& Discussion}

\subsection{Mechanical characterization}
\begin{figure}
    \centering
    \includegraphics[width=1\linewidth]{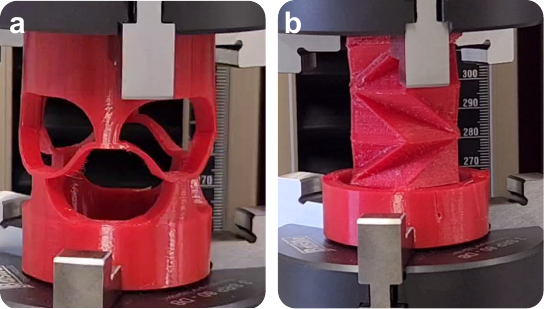}
    \caption{Mechanical testing of the two separated structures: a) Metashell. b) Kresling origami.}
    \label{mechanical}
\end{figure}

\begin{figure}
    \centering
    \includegraphics[width=0.9\linewidth]{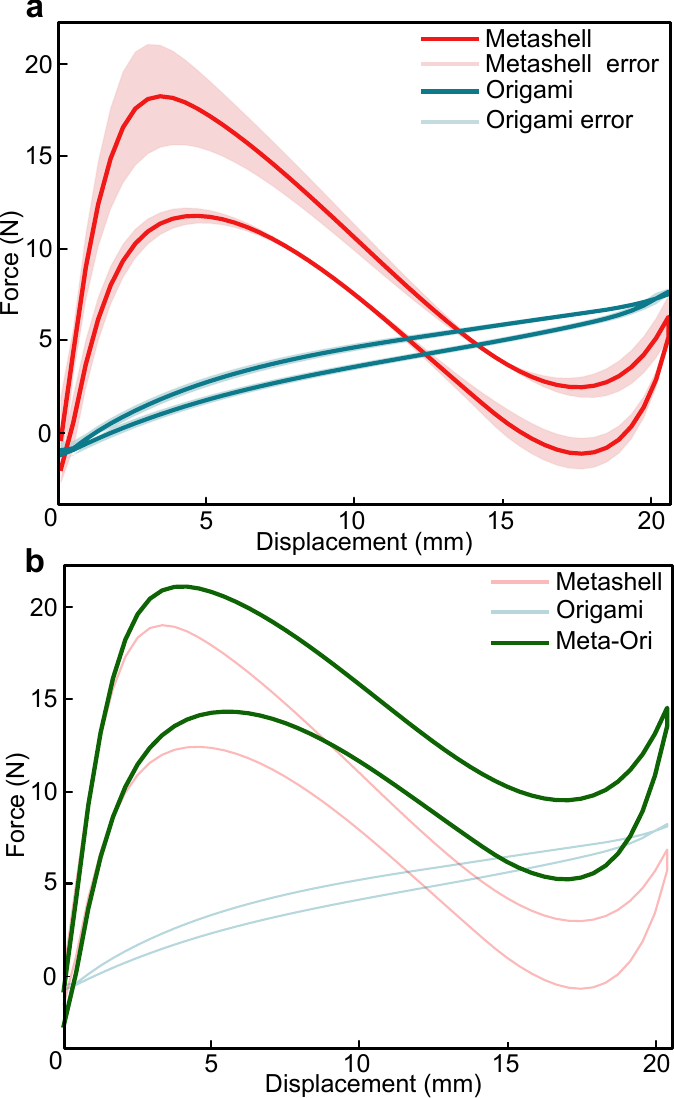}
    \caption{a) Mechanical response of the metashell and the origami. b) Virtual stiffness of the Meta-Ori by summing the two measured mechanical responses.}
    \label{mechtest}
\end{figure}

We characterize the quasi-static response of the metashell and the origami separately (Fig. \ref{mechanical}). Fig. \ref{mechtest} depicts the force-displacement curves for the two structures averaged over the cycles. The metashell is initially in its open state, which is the printed configuration, which means that the structure is first compressed and subsequently pulled. The metashell curve has a clear nonmonotonic behavior featuring snapping points and bistability as the curve crosses the zero force point. The origami structure is also tested with the same loading condition. As expected, the origami has a monotonic response with a lower stiffness compared to the metashell. As explained in the previous section, the final Meta-Ori stiffness can be thought of as the summing contributions of the metashell and the origami, where we want the nonmonotonic behavior to be preserved. As such, we plot the "virtual stiffness" curve of a Meta-Ori by summing the measured curves of the two different structures in Fig. \ref{mechtest}b. Interestingly, the summed force-displacement curve is nonmonotonic as the origami stiffness is not high enough to linearize the global response, and the snapping points are dictated by the metashell.

\begin{figure*}[ht!]
    \centering
    \includegraphics[width=.7\linewidth]{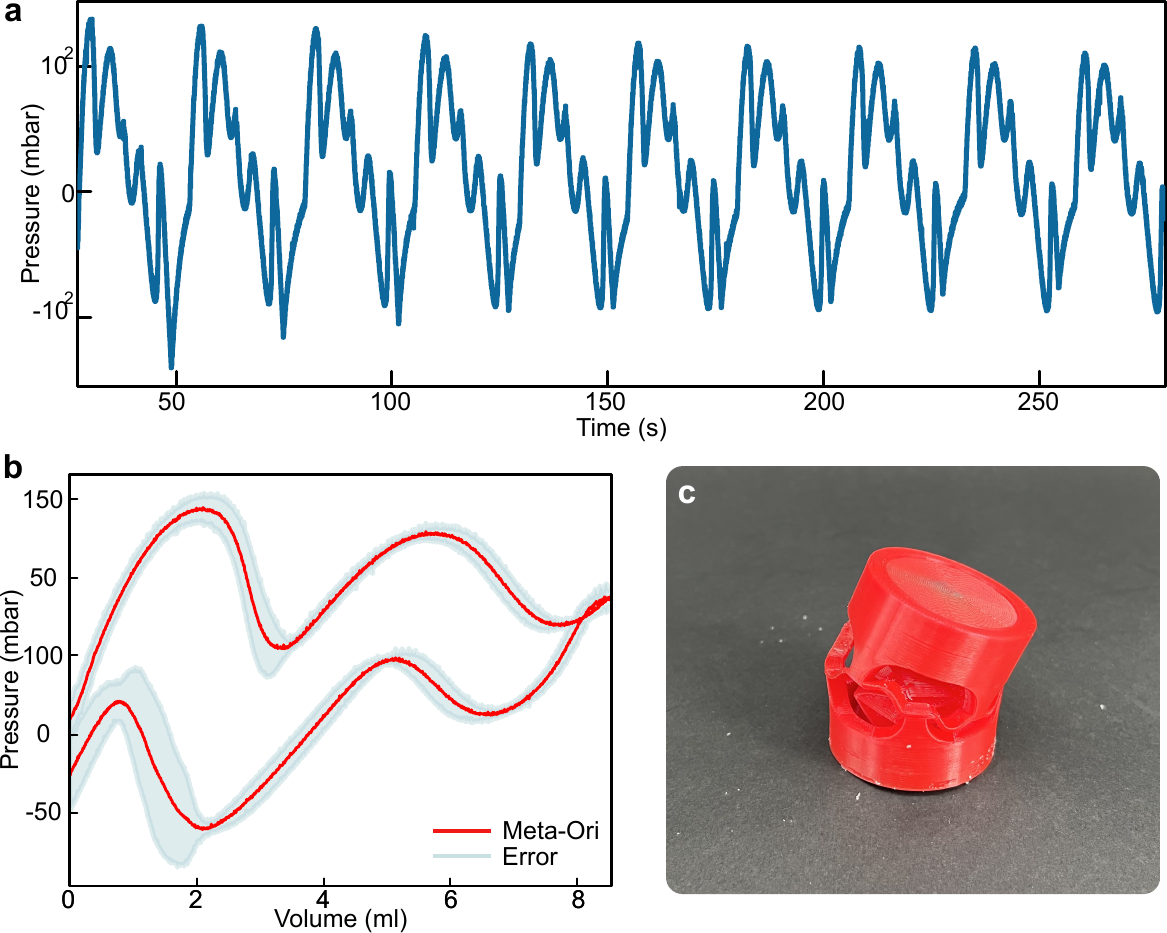}
    \caption{Fluidic characterization of the Meta-Ori. a) Pressure variation in time over 10 cycles and b) average PV curve over one cycle. Negative pressure values show that the actuator is bistable. c) Intermediate snapping state of the Meta-Ori.}
    \label{pv}
\end{figure*}

\subsection{Pressure-Volume curves}
The pressure-volume relation of the Meta-Ori is displayed in Fig. \ref{pv}. The curve is highly nonlinear and features two nonlinear pressure peaks in inflation and two in deflation, which correspond to multiple snapping transitions. The Meta-Ori is a bistable inflatable as negative pressure is generated during deflation, which means that the actuator needs vacuum to be pulled back in its second state. Fig. \ref{pv}a depicts the 10 cycles of inflation and deflation, showing the consistency of the snapping phenomena. The two snapping events in inflation and deflation are caused by the asymmetric snapping of the metashell. The first intermediate snap is caused by one unit-cell which opens before the other three, so that the specular unit-cell is still closed and the two neighboring ones are partially open. As a result, the structure tilts to one side (Fig. \ref{pv} c). As the inflation continues, the other unit-cells snap open and the structure re-aligns and stays in this stable state. This asymmetry does not happen in the mechanical tests described in the previous subsection because the metashell is uniformly clamped to the testing machine and no bending motion is allowed (Fig. \ref{mechanical}a). It is interesting to notice that the pressure required for our Meta-Ori inflatable is in the range of 150 mbar, despite being made out of a relatively stiff elastomer (Shore A95). For comparison, highly-optimized PneuNet-like printed inflatable actuators require more than 230 mbar even when made out of softer elastomers (Shore A80) \cite{zhai2023desktop}. This is thanks to the foldable deformation mechanism of the origami as opposed to the mostly stretching mechanism of the PneuNets.

\subsection{Actuation sequencing}
\begin{figure}[t!]
    \centering
    \includegraphics[width=1\linewidth]{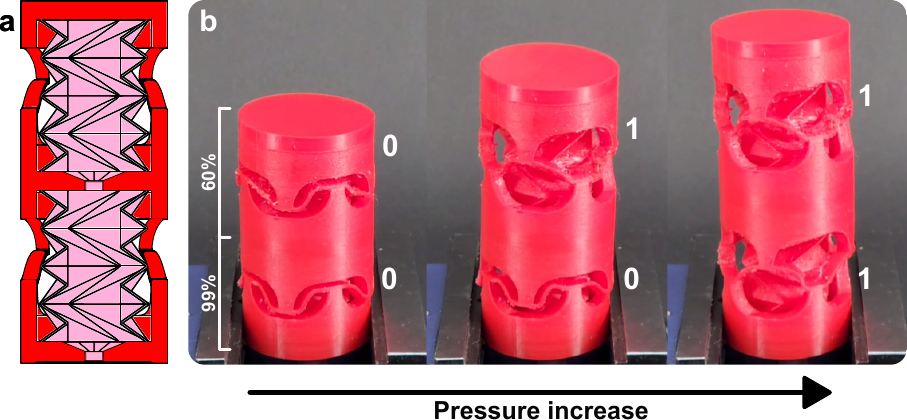}
    \caption{a) Cross-section view of the bi-segment Meta-Ori. b)Snapping sequence with segments having 60\% (top) and 99\% (bottom) of infill}
    \label{sequence}
\end{figure}

A typical demonstration of the functionalities of nonlinear snapping actuators is the pre-programmed discrete sequences of motion upon actuation. Actuation sequencing can be used in locomotion \cite{gorissen2019hardware}, automated music players \cite{van2023nonlinear}, traveling waves \cite{milana2022morphological}, and consists of using only a single input to control multiple outputs by harnessing the snap-through events as intrinsic mechanical switches. We fabricate a bi-segment monolithic Meta-Ori which has a metashell with a 2x4 unit-cell arrangement (two rows with four circumferential unit-cells) and an inflatable origami along the whole length, as depicted by the cross-section view of Fig. \ref{sequence}a. Each row corresponds to one segment. For sequencing to occur, the snapping points of the segments need to differ, so that by controlling the pressure threshold it is possible to trigger the snapping events in a discrete sequence. We differentiate the two segments by changing the amount of infill of the two rows of the metashell, where the bottom one has 99\% infill and the top one has 60\% infill. The segment with higher infill has stiffer support beams and walls, thus requiring more pressure to snap, compared to the segment with the lower infill. As depicted in Fig. \ref{sequence}b and shown in the supplemental video, as the pressure increases the top segment snaps first followed by the second one.

\section{Conclusions}

This work introduces the Meta-Ori, a novel monolithic nonlinear inflatable actuator combining nonlinear cylidrical mechanical metamaterials - metashells - with Kresling origami inflatables to achieve complex and tunable nonlinear mechanical behaviors and increase the physical functionalities of soft robotic systems. Our approach enables new snap-through actuators with a monolithic design that can be fully 3D printed, avoiding additional manual assembly steps and enhancing structural integrity. We characterize the mechanics of our metashell design and the origami, showing that the nonlinear mechanics of the Meta-Ori depends on the metashell characteristics, where the origami acts as a pressure transmitter. Additionally, we demonstrate actuation sequencing using a single input in a bi-segment monolithic Meta-Ori, where the sequencing depends on the different stiffness of the two segments, caused by different infill percentages assigned during the printing process. Lastly, we publicly share the Grasshopper visual code developed in this work to provide the soft robotics community with an accessible tool for designing origami fluidic actuators, aiming to democratize and promote this technology. The code is available at \texttt{https://www.softmachineslab.org/software}.

The results demonstrate that Meta-Ori actuators exhibit a highly nonlinear pressure-volume relationship and consistent snapping behavior. The actuator’s ability to sequentially snap in multi-segment configurations showcases potential soft robotic applications where discrete, programmable actuation is desired, such as locomotion, sequential deployment or grasping.

In conclusion, our work shows a novel design that harnesses the mechanical programmability of metamaterials to augment soft actuators functionality. Our Meta-Ori has an elongation of 43\% from the closed to the open state, improving the stroke performance compared to other nonlinear soft actuators - balloons, membranes, and multistable origami. Future work will explore new geometries and architected arrangements of the unit-cells of Meta-Ori to have multistable states for shape morphing or complex sequencing for physically intelligent soft robots. Further, we will implement conductive soft materials to integrate the Meta-Ori with self-sensing capabilities.


\section*{Acknowledgment}
This work was funded by the Deutsche Forschungsgemeinschaft (DFG, German Research Foundation) under Germany’s Excellence Strategy-EXC-2193/1-390951807.

\bibliographystyle{ieeetr}
\bibliography{references}
\end{document}